\documentclass[10pt, onecolumn]{article}   	
\usepackage{geometry}                		
\pdfoutput=1                 		
\usepackage{graphicx}
\usepackage{subfigure}				
\usepackage{amssymb}
\usepackage{url}
\title{\bf The $\mathcal{E}$-Average Common Submatrix: Approximate Searching in a Restricted Neighborhood}
\author{Alessia Amelio$^1$\footnote{(Corresponding author)}, Darko Brodi\'c$^2$}
\date{\small $^1$ Department of Computer Engineering, Modeling, Electronics and Systems, University of Calabria, \\Via P. Bucci Cube 44, 87036 Rende (CS), Italy,\\aamelio@dimes.unical.it\\$^2$ Technical Faculty in Bor, University of Belgrade, Vojske Jugoslavije 12, 19210 Bor, Serbia, \\dbrodic@tfbor.bg.ac.rs}							

\begin{document}
\maketitle

\begin{@twocolumnfalse}

{\bf Abstract}.
This paper introduces a new (dis)similarity measure for 2D arrays, extending the Average Common Submatrix measure. This is accomplished by: (i) considering the frequency of matching patterns, (ii) restricting the pattern matching to a fixed-size neighborhood, and (iii) computing a distance-based approximate matching. This will achieve better performances with low execution time and larger information retrieval.
\end{@twocolumnfalse}
\vspace{1cm}

\section{Introduction}
A similarity measure quantifies how similar two objects, or parts of objects, are. 
Basically, the use of certain similarity metrics deeply depends on the problem to be solved. In particular, it defines what is the goal of such a measuring process, i.e. the level of matching target and reference object.
Metrics heavily depend on the characteristics of the objects to be compared. Typically, they are: (i) mean squares, (ii) normalized correlation, (iii) pattern intensity, (iv) mutual information, and (v) self-similarity matrix. 
Mean squares, which represent the sum of squared differences between intensity values, use these intensity values given by two objects, which should be in the same range. 
Normalized correlation calculates the correlation between the intensity values that characterize the two objects, which can establish a linear relationship.
Pattern intensity squares the differences between intensity values transformed by a function, which are at the end summed. Mutual information is a metrics extracted from information theory, based on the concept of entropy. Self-similarity matrix graphically represents similar sequences in a specific data series. 

In this paper, we propose a new (dis)similarity measure for 2D arrays which can be used in different contexts, including images, documents and sensor data. The paper is organized as follows. Section 2 introduces the Average Common Submatrix measure from which the new (dis)similarity measure derives. Section 3 presents the main features of the new (dis)similarity measure. Section 4 briefly describes the experimental setting. At the end, Section 5 draws a conclusion. 

\section{The Average Common Submatrix}
Let {\bf A} and {\bf B} be two square matrices respectively of area $n \times n$ and $m \times m$, defined on the same alphabet $\Sigma$. The Average Common Submatrix measure (ACSM) \footnote{This work was realized at Georgia Institute of Technology, GA, USA (Caianiello Best Young Scientist Paper Award 2013)} 
computes the average area of the largest square sub-matrices in common in {\bf A} and {\bf B} to quantify the similarity between {\bf A} and {\bf B} \cite{[1]}. Let {\bf C} be another square matrix of size $k \times k$ defined on the same alphabet $\Sigma$. Then, {\bf A} will be more similar to {\bf B} than to {\bf C} if the average area of the sub-matrices shared between {\bf A} and {\bf B} is higher than that between {\bf A} and {\bf C}.

The ACSM similarity is defined as follows:
\begin{equation}
S_\alpha({\bf A},{\bf B}) = \frac{\sum_{i,j=1}^n W(i,j)}{n^2}, \hspace{0.5cm}W(i,j)\ge \alpha
\end{equation}
where $\alpha$ is a parameter setting the minimum area of the sub-matrices to consider.

For each position $(i,j)$ in {\bf A}, the largest square sub-matrix starting at that position exactly matching with a sub-matrix at some position $(k,h)$ in {\bf B} is found. This is accomplished by considering the sub-matrix of maximal area $min\{i,j\} \times min\{i,j\}$ starting at position $(i,j)$ in {\bf A} and trying to find a match inside {\bf B}. If any match occurs, the square sub-matrix of smaller area $(min\{i,j\}-1) \times (min\{i,j\}-1)$ at position $(i,j)$ in {\bf A} is considered and an exact match is searched again inside {\bf B}. This procedure is repeated until the considered square sub-matrices have an area $\ge \alpha$ or a square sub-matrix which exactly matches inside {\bf B} is found. This is the largest square sub-matrix at position $(i,j)$ of area $W(i,j)$. Exact match imposes that all elements of the sub-matrices have to be identical at the corresponding positions.
From the similarity measure, the dissimilarity measure is then computed (see in Ref. \cite{[1]}).
 
The brute-force algorithm for computing the ACSM measure takes $O(m^2n^3)$ time. In fact, for each of the $n^2$ positions in {\bf A} a maximum of $n$ sub-matrices is searched in {\bf B}, and finding the exact match of a sub-matrix  in {\bf B} takes $O(m^2)$ \cite{[5]}. Although a solution using a generalized suffix-tree has been introduced \cite{[2]}, the construction of the tree remains a quite demanding task for large images. 

Furthermore, when the size of the alphabet $\Sigma$ is particularly high, the matrices {\bf A} and {\bf B} will have a high variability in their elements. Because of the exact match, this may compromise the dissimilarity computation, resulting in quite high values for small effective variations. To overcome this limitation, an approximate version of ACSM measure has been introduced, omitting the contribution of some elements at regular intervals in the match of the sub-matrices \cite{[3]}. However, setting the size of the interval is a critical point which may result in information loss during the computation of the dissimilarity. 

\section{The $\mathcal{E}$-Average Common Submatrix}
In the following, we introduce the $\mathcal{E}$-Average Common Submatrix ($\mathcal{E}$-ACSM), which is an approximate and faster version of ACSM. 
The $\mathcal{E}$-ACSM is framed into two axiomatic conditions, providing a theoretical background on which the measure has been defined.
In particular, the first axiomatic condition proposes to consider also the number of positions in {\bf A} with a sub-matrix matching inside {\bf B} to quantify the similarity between {\bf A} and {\bf B}, which is a brand new idea. The second axiomatic condition formalizes the concept of similarity as currently defined in ACSM.
This is the starting point to propose a new distance-based approximation strategy where no elements are omitted in the match of the sub-matrices. It is added to a new matching strategy providing a clear advantage in execution time. In fact, matching of the sub-matrices of {\bf A} is only performed in a restricted neighborhood inside {\bf B}.
In most cases, this matching strategy is enough to guarantee a reliable (dis)similarity computation in a reduced time.

\subsection{Axiomatic Conditions}
The two axiomatic conditions provide a variant in 2D of the axiomatic conditions introduced by Ziv in \cite{[4]} for the sequence similarity. 
In particular, let $p_1$ be the fraction of positions in {\bf A} for which a sub-matrix matching inside {\bf B} exists. Also, let $p_2$ be the fraction of positions in {\bf A} corresponding to matching of distinct sub-matrices in {\bf B}: (i) matrix {\bf B} will be considered dissimilar to matrix {\bf A} if $min\{p_1,p_2\} < p_0$, where $p_0$ is a parameter, (ii) let {\bf B} and {\bf C} be two matrices which are considered similar to {\bf A} according to the first 
axiomatic condition. Then, {\bf B} and  {\bf A} will be more similar to each other than {\bf B} and {\bf C} if the average area of the square sub-matrices in common between {\bf A} and {\bf B} is higher than the same average area between {\bf A} and {\bf C}.

The first condition examines the number of positions in {\bf A} with a sub-matrix matching inside {\bf B}, i.e. the frequency of the square sub-matrices matching in {\bf A} and {\bf B}. Also, it imposes that sub-matrices of {\bf A} at different positions have to match to a given number of distinct sub-matrices in {\bf B}. Hence, if there is a sub-matrix for each position $(i,j)$ in {\bf A} matching with a single sub-matrix inside {\bf B}, {\bf A} and {\bf B} will be dissimilar.
The second condition determines the current methodology for ACSM computation.

\subsection{Restricting the Neighborhood}
After introducing the frequency of matching the sub-matrices, we impose stricter constraints in the matching procedure. In particular, for each position $(i,j)$ in {\bf A} we find the largest square sub-matrix matching at some position $(k,h)$ in a $\mathcal{E}$-neighborhood of the same position $(i,j)$ in {\bf B}. Hence, we restrict the matching inside {\bf B} to an area of size $\mathcal{E} \times \mathcal{E}$ around $(i,j)$, where $\mathcal{E}$ is a constant parameter. 

This determines an improvement of the ACSM execution time, with an impact on the complexity of the algorithm. Recall that matching of a sub-matrix of {\bf A} inside {\bf B} is linear in the area of {\bf B}, taking $O(m^2)$ \cite{[5]}. In this case, only a restricted area of fixed size $\mathcal{E} \times \mathcal{E}$ is considered for matching, where $\mathcal{E} << m$. Consequently, matching of a sub-matrix of {\bf A} inside {\bf B} will be reduced to $O(\mathcal{E}^2)$.

\subsection{Distance for Approximate Matching}
To avoid omitting some elements during the match of the sub-matrices \cite{[3]}, we compute a distance function between them. Hence, the exact match of the elements at the corresponding positions will be substituted by distance computation between the sub-matrices. Then, a positive match will be obtained between the sub-matrices if their distance is lower than a threshold parameter $\tau$. 

Formally, let ${\bf A}(i,j)$ and ${\bf B}(k,h)$ be two square sub-matrices at position $(i,j)$ of {\bf A} and $(k,h)$ of {\bf B} in the $\mathcal{E}$-neighborhood of $(i,j)$. The two sub-matrices will have a positive match if:
$d(\overline{{\bf A}}(i,j),\overline{{\bf B}}(k,h)) < \tau$,
where  $\overline{{\bf A}}(i,j)$ and $\overline{{\bf B}}(k,h)$ are the sub-matrices in some feature space, and $d: X \times X \to [0, \infty)$ is a distance function over the elements $X$ of the feature space.

\section{Experimental Setting}

We expect $\mathcal{E}$-ACSM to obtain good performances on synthetic and real-world databases \cite{[1]}, \cite{[2]}, with a lower execution time than ACSM, and a more accurate information retrieval than the approximate ACSM \cite{[3]}. Also, embedding the frequency information of the common sub-matrices will provide a more reliable similarity evaluation than ACSM.

\section{Conclusions}
This paper presented $\mathcal{E}$-ACSM, a new (dis)similarity measure extending the baseline ACSM in three aspects: (i) considering the frequency of the shared sub-matrices, and not only their area, in the similarity computation, (ii) restricting the matching in a neighborhood of the position, and (iii) realizing the approximate matching by a distance function. 
Setting the neighborhood size and the distance function makes $\mathcal{E}$-ACSM more flexible than ACSM to be employed in different contexts. In the future, we are planning to experiment $\mathcal{E}$-ACSM in three domains: (i) images, (ii) digitized documents, and (iii) 2D data representations generated by sensors for which a (dis)similarity computation is required.

\end{document}